\documentclass{ieeeaccess}
\usepackage{cite}
\usepackage{amsmath,amssymb,amsfonts}
\usepackage{graphicx}
\usepackage{textcomp}
\usepackage{lipsum}  
\usepackage[normalem]{ulem}
\usepackage{stfloats}
\usepackage{breqn}
\usepackage{hyperref}
\hypersetup{
    colorlinks = true,
    linkbordercolor = {},
    allcolors  = blue
}

\usepackage{epstopdf}
\usepackage[caption=false]{subfig}
\usepackage{amsmath}
\usepackage{soul,color}
\usepackage{hyperref}
\hypersetup{
    colorlinks = true,
    linkbordercolor = {},
    allcolors  = blue
}

\usepackage{adjustbox,graphicx}
\usepackage{ctable}
\usepackage{changepage}
\usepackage{xcolor}
\usepackage{colortbl}
\makeatletter
\patchcmd{\@makecaption}
  {\scshape}
  {}
  {}
  {}
\makeatletter
\patchcmd{\@makecaption}
  {\\}
  {.\ }
  {}
  {}
\makeatother
\def\tablename{Table}

\newcommand\tab[1][1cm]{\hspace*{#1}}   
\newcommand{\minA}{\operatornamewithlimits{min}} 

\usepackage{algorithm}
\usepackage{algpseudocode}

\def\BibTeX{{\rm B\kern-.05em{\sc i\kern-.025em b}\kern-.08em
    T\kern-.1667em\lower.7ex\hbox{E}\kern-.125emX}}
\begin{document}

\history{Date of publication xxxx 00, 0000, date of current version xxxx 00, 0000.}
\doi{10.1109/ACCESS.2017.DOI}

\title{{
A Functional Model for Structure Learning and Parameter Estimation in Continuous Time Bayesian Network: An Application in Identifying Patterns of Multiple Chronic Conditions}
}
\author{\uppercase{Syed Hasib Akhter~Faruqui}\authorrefmark{1}, 
\uppercase{Adel~Alaeddini\authorrefmark{1}, Jing~Wang\authorrefmark{2}, 
Carlos~A.~Jaramillo\authorrefmark{3},
and Mary~Jo~Pugh\authorrefmark{4}
},
}

\address[1]{Department of Mechanical Engineering, The University of Texas at San Antonio, San Antonio, USA. (e-mail:  \{syed-hasib-akhter.faruqui, adel.alaeddini\}@utsa.edu)}
\address[2]{School of Nursing, UT Health San Antonio, San Antonio, TX, USA.}
\address[3]{South Texas Veterans Health Care System, San Antonio, TX, USA.}
\address[4]{VA Salt Lake City Health Care System, Salt Lake City, UT, USA.}

\tfootnote{This research work was supported by the National Institute of General Medical Sciences of the National Institutes of Health under award number "1SC2GM118266-01".}


\corresp{Corresponding author: Adel Alaeddini (e-mail: \url{adel.alaeddini@utsa.edu}).}

\begin{abstract}
Bayesian networks are powerful statistical models to study the probabilistic relationships among sets of random variables with significant applications in disease modeling and prediction. Here, we propose a continuous time Bayesian network with conditional dependencies, represented as regularized Poisson regression, to model the impact of exogenous variables on the conditional intensities of the network. We also propose an adaptive group regularization method with an intuitive early stopping feature based on Gaussian mixture model clustering for efficient learning of the structure and parameters of the proposed network. Using a dataset of patients with multiple chronic conditions extracted from electronic health records of the Department of Veterans Affairs, we compare the performance of the proposed approach with some of the existing methods in the literature for both short-term (one-year ahead) and long-term (multi-year ahead) predictions. The proposed system provides a sparse intuitive representation of the complex functional relationships between multiple chronic conditions. It also provides the capability of analyzing multiple disease trajectories over time, given any combination of preexisting conditions.
\end{abstract}
\begin{keywords} 
Continuous Time Bayesian Network, Poisson Regression, Adaptive Group Lasso, Gaussian Mixture Model, Multiple Chronic Conditions.
\end{keywords}
\titlepgskip=-15pt

\maketitle
\section{Introduction}
\label{Section:Introduction}

\IEEEPARstart{B}{ayesian} networks (BNs) are probabilistic graphical models that represent a set of random variables and their conditional dependencies via a directed acyclic graph (DAG) \cite{pearl_theory_1995, heckerman_tutorial_1998, yang2016learning}. BNs offer valuable insights about the random variables and their interactions for complex data summarization and visualization, prediction and inference, and correlation and causation analysis by encoding the information uncertainty in their structure. BNs also have useful applications in medicine for predictive modeling of multiple chronic conditions (MCC) \cite{faruqui2018mining, lappenschaar2013multilevel}.

Although BNs were originally designed for studying the relationships among static random variables, recently, it has been applied to study random variables with temporal behavior \cite{kanazawa1991logic, aliferis1996structurally, lappenschaar2013multilevel}. 
Multilevel temporal Bayesian networks  (MTBNs) describe the temporal states of the network variables over a finite number of discretized times. \cite{lappenschaar2013multilevel, faruqui2018mining}.
The set of edges within each discretized time present the regular conditional dependencies among random variables, while the edges between the (discretized) time points represent the temporal dependencies. Since MTBNs do not directly model the time and the dynamics of the random variables, classic structure learning algorithms can be used to learn the structure and parameters of the network.
Dynamic Bayesian networks (DBNs) \cite{kevin_dynamic_2002, dean_model_1989,dagum1992dynamic} are another extension to BNs, that represent the temporal dynamics of random variables over an infinite number of discretized times.
Unlike MTBNs, DBNs generally duplicate the time slices to represent the temporal dynamics of the random variables over a fixed time range and do not allow for a change in the structure of the network over time\cite{arroyo1999temporal}. 
Temporal Nodes Bayesian networks (TNBNs) are yet another alternative for modeling the dynamic processes of BNs random variables. The nodes of TNBNs represent the time of occurrence, and the edges represent the causal-temporal relationships. The temporal nodes allow for having time intervals of different durations to represent the possible delays between the occurrences of parent events (causes) and the corresponding child events \cite{arroyo2005temporal}.

MTBN{s}, DBN{s}, and TNBN{s} describe the states of temporal BNs over discrete time points but do not model time explicitly. This makes it very difficult to query MTBNs, DBNs, and TNBNs over the time at which the state of a random variable change or an event occurs (i.e., at an irregular time). Furthermore, MTBN{s}, DBN{s}, and TNBN{s} slice the time into fixed increments, but in reality, the random variables such as chronic conditions evolve at different time granularities. This makes the inference process very challenging, especially for large-scale networks. Choosing a large or small granularity may change the network structure and cause inaccurate model (for {a} large time granularity) and learning/inference inefficiencies (for {a} small-time granularity) {\cite{nodelman2002learning, chatterjee2010dbns, wang2012continuous}}.

Continuous time Bayesian networks (CTBNs) \cite{nodelman_continuous_2002}, on the other hand, explicitly model the time by defining a graphical structure over continuous time Markov processes (CTMPs)\cite{el2012continuous}. This allows explicit representation of the temporal dynamics and the probability distribution of the random variables over time, i.e., the emergence of a new chronic condition in MCC patients. 
However, CTBNs assume fixed conditional intensities for representing the relationships between random variables and, therefore, cannot model the impact of exogenous variables on the conditional dependencies of the network. 
Additionally, similar to DBN{s}, TNBN{s} and, MTBN{s}, learning the structure of CTBN{s} is challenging and typically carried out by heuristic greedy search algorithms \cite{chickering2004large}. This restricts the application of CTBN{s} to problems with {multiple exogenous variables of different levels}.
An example of this problem is when modeling the temporal relationship between the emergence of different chronic medical conditions which is affected by individual patient{s'} gender, age, race, education{,} etc.

To address the above challenges, we propose to represent the conditional intensities (dependencies) of the CTBN as {regularized} Poisson regression to take into account the impact of various levels of exogenous variables on the network structure and parameters. We then transform the CTBN into a large-scale regularized regression estimation problem and propose an adaptive framework with early stopping features for structure and parameter learning. Using a large dataset of patients with multiple chronic conditions extracted from electronic health records of the Department of Veterans Affairs, we compare the predictive performance of the proposed functional CTBN model with some of the existing methods in the literature, including LRMCL \cite{alaeddini_mining_2017} and unsupervised MTBN \cite{faruqui2018mining}. We also demonstrate the performance of the proposed functional CTBN for analyzing the trajectories of MCC emergence over time.
{
Our paper has the following contributions:
\begin{enumerate}
    \item We propose to formulate the conditional intensities of the continuous-time Bayesian networks (CTBN) as a function of exogenous risk factors using regularized Poisson regression. Our case study enables the personalization of the proposed functional CTBN for individual patients according to their risk factors.
    \item We propose an adaptive group regularization-based framework to simultaneously learn the structure and conditional intensities of the functional CTBN. The information of the regularization path of the proposed learning algorithm helps the users, i.e., medical practitioners and patients, to achieve the desired level of sparsity.
    \item We propose a Gaussian mixture model (GMM) based approach for early stopping of the proposed learning algorithm without losing much information. The proposed approach uses clustering to expedite pushing insignificant parameters with very small values toward zero, which may take numerous additional iterations of the training algorithm. 
\end{enumerate}
}

The remainder of the paper is structured as follows. Section \ref{Section: Literature} provides the relevant literature to the proposed study. Section \ref{Section: background} presents the preliminaries and background for the CTBN. Section \ref{Section: Methodology} describes the details of the proposed functional CTBN and the regularized regression model for learning its structure and parameters. Section \ref{Section:CTBN_Case_Study} presents the study population, the resulting model structure and parameters, predictive performance, and trajectory analysis. Section \ref{Section: Conclusion} provides the summary and concluding remarks.

\vspace{-1.0em}
\section{Relevant Literature}
\label{Section: Literature}
CTBN{s} are graphical model{s} whose nodes are associated with random variables with states continuously evolving over time. Consequently, the evolution of each variable depends on the state of its parents in the graph. Nodelman et al \cite{nodelman_continuous_2002, nodelman2002learning} presented the framework of CTBN in their previous work{s}. It was built on the framework of homogeneous Markov processes \cite{norris1998markov}, which provided the model of evolution in continuous time and at the same time utilizing the ideas of Bayesian networks to provide a graphical representation for a system. CTBNs overcome the limitations of other temporal models ({MTBNS}, DBN{s}, TNBN{s}, etc.) by explicitly representing temporal dynamics of a system i.e. {they} can learn the probability distribution over time for systems (processes) that evolve at an irregular time interval \cite{nodelman_continuous_2002}. CTBN{s} have been used for a variety of dynamic temporal systems like discovering the social network dynamics \cite{fan2012learning}, intrusion in network computer system \cite{xu2010intrusion}, modelling sensor networks \cite{shi2010intelligent}, reliability analysis of dynamic systems \cite{boudali2006continuous}, robot motion monitoring \cite{ng2005continuous}, and monitoring and predicting cardiogenic heart failure \cite{gatti2012continuous}.

Nodelman et al. \cite{nodelman2002learning} derived a Bayesian scoring function to learn the structure of a CTBN model from fully observed data. However, in real life, we often obtain partially observable data. Thus later, they provided an extension to learn the structure of {a} CTBN from partially observable data based on the structural EM algorithm \cite{nodelman2012expectation2}. Codecasa et al. \cite{codecasa2013conditional} extended the CTBN structure learning model presented by Nodelman et al \cite{nodelman2002learning} to {a} CTBN classifiers by constraining the class nodes (not dependent on time). Their model combines conditional log-likelihood scoring with Bayesian parameter learning, which outperformed the previous log-likelihood scoring function. Yang et al. \cite{yang2016learning} developed a non-parametric approach to learn {a} CTBN structure in relational domains, with varying numbers of objects and {the} relations among them.  

Although a CTBN provides a compact representation over traditional CTMP, for a large or highly inter-dependent system, the complexity {of} learning a CTBN model grows exponentially with respect to a node's parents. In the worst case, a node may depend on all other nodes in the network, resulting in a complexity equivalent to the original CTMP. 

Perreault et al. \cite{perreault2019compact} imposed additional structures on the model to reduce the complexity in learning the CTBN models. 
Cao \cite{cao_2011} modeled a logical OR gate utilizing the CTBN nodes with deterministic transitions. Logan et al. \cite{logan_2016} extended the Noisy-OR for CTBNs to reduce the required parameters. In this paper, to reduce the number of parameters to estimate, we assume the conditional effects of the parent’s nodes are multiplicative, which is on a par with the Noisy-OR \cite{pearl_1988_reasoning, Onisko_2001} and the CT-NOR \cite{simma_2008}. The Noisy-OR model assumes the independence of the effects of parent nodes to reduce the model complexity. Natural parameterization of the NOR model is equivalent to the CT-NOR model in the limit, given the bin width approaches zero.

The inference process in CTBN{s} is different than inference in general BN models. Both the exact inference and approximate inference in CTBN{s} are \textit{NP}-hard even if the initial state values are given \cite{sturlaugson2014inference}. The exact inference \cite{nodelman2002learning} in CTBN{s} utilizes the full joint intensity matrix and computes the exponential of the matrix, which is often intractable. This method of inference often ignores the factored nature of the CTBNs; thus, most research in CTBNs' inference has focused on approximation algorithms \cite{fan2010importance}. Nodelman et al. \cite{nodelman2012expectation} developed such an approximation inference method based on expectation propagation. Later, Saria et al. \cite{saria2012reasoning} extended the model to full belief propagation and provided an algorithm to adapt the approximation quality. A message-passing scheme has been employed in neighboring nodes for each interval of evidence provided. Messages are continually passed till a consistent distribution has been attained over the interval of evidence. Several sample based algorithms have also been developed. El-hay et al. \cite{el2012gibbs} developed a Gibbs sampling based procedure to sample from the trajectories given a certain set of parent conditions while Fan et al. \cite{fan2008sampling} developed an importance sampling algorithm that computes the expectations of any function of trajectory to perform the inference operation given a fixed set of constraints. Methods using variational techniques such as {the} belief propagation \cite{el2010continuous} and {the} mean-field approximation \cite{cohn2010mean} have also been developed. These models utilize systems of ordinary differential equations to approximate the system distribution. To handle point evidence, Ng et al. \cite{ng2005continuous} developed a continuous time particle filtering algorithm.

Aside from CTBNs, a significant amount of work has also been done to integrate Poisson processes with Bayes nets to represent events in continuous time. Rajaram et al. \cite{rajaram_2005} developed the Poisson network model for representing multivariate structured Poisson processes. They modeled the waiting times of a process by an exponential distribution with a piecewise constant rate function that depends on the event counts of its parents. They also adopt a Bayesian approach for learning the network structure. Simma et al. \cite{simma_2008} presented CT-NOR, a generative model for representing and reasoning about the relationships among events in continuous time. Using a parameterized function, the CT-NOR can incorporate specific domain knowledge about the expected shape of the distribution of the time delay between events. They used the expectation-maximization (EM) algorithm to fit the parameters of the CT-NOR model. Simma et al. \cite{simma_2010} presented a framework for building a probabilistic model of discrete events over continuous time based on cascades of Poisson processes. Their Poisson cascades model can exploit a wide range of delays, transitions, and fertilities. They used the EM algorithm to inference from the Poisson Cascades model. Gunawardana et al. \cite{gunawardana_2016} described a set of graphical event models (GEMs) to approximate a board class smooth multivariate temporal point processes. They used BIC and ML for parameter and structure learning. They also provided theoretical results showing that the dependency structure of a universal family of point process models can be learned from data.

In this paper, we extend the earlier works in the literature by formulating the conditional intensities of the transitions between the states of the CTBN as a regularized Poisson regression of exogenous risk factors, while assuming the multiplicative effect of parent nodes to reduce the number parameters to estimate. We also propose using principal component analysis (PCA) or kernel PCA to extract few informative features of exogenous risk factors and reduce the dimensionality. We develop an adaptive group regularized regression-based framework to simultaneously learn the structure and parameters of the proposed functional CTBN model, where the information of the regularization path of the learning algorithm allows for achieving the desired level of sparsity. Additionally, we represent a Gaussian mixture model to enable early stopping of the estimation procedure.
\vspace{-1.0em}
\section{Relevant Background}
\label{Section: background}
In this section we review major components of {a} CTBN \cite{nodelman_continuous_2002, nodelman2002learning}. {A} CTBN represents finite-state, continuous-time processes over a factored state, which explicitly represents the temporal dynamics and allows to extract the probability distribution overtime when a specific event occurs. 
\vspace{-1.0em}
\subsection{Markov Process}
\label{Subsection: markov_process}
{Markov processes are} an important class of random processes in which the future state of a random variable is independent of the past, given the present \cite{el2012continuous}.
Let $\textbf{X} = \{x_1,x_2,....,x_n\}$ denotes the state space of the random variable $X$. The stochastic behavior of $X$ can be modeled by an initial distribution $P_X^0$ and a time-invariant transition intensity matrix $Q_X$ of size $n\times n$ which can be written as 

\[
\textbf{Q}_{X} = \begin{bmatrix} 
    -q_{x_1}    & q_{x_1x_2}    & \dots & q_{x_1x_n} \\
    q_{x_2x_1}  & -q_{x_2}      & \dots & q_{x_2x_n} \\
    \vdots      & \vdots        & \ddots & \vdots\\
    q_{x_nx_1}  & q_{x_nx_2}    & \dots & -q_{x_n}
    \end{bmatrix}
\]

\noindent where $q_{x_ix_j}$ represents the intensity of the transition from state $x_i$ to state $x_j$, and $q_{x_i} = \sum_{j \neq i} q_{x_ix_j}$. 
The probability density function ($f$) and the probability distribution function ($F$) for staying at the same state (say, $x_i$), which is exponentially distributed with parameter $q_{x_i}$, are calculated as-

\begin{align}
\label{Equation:Probability density}
    f(q_x,t) &= q_{x_i} exp(-q_{x_i}t),&{t \geq 0}\\
    F(q_x,t) &= 1 - exp(-q_{x_i}t),    &{t \geq 0}
\end{align}

\noindent 
After transitioning, which takes an expected transition time of $\frac{1}{q_{x_i}}$, the variable $X$ shifts to state $x_j$ with probability $\theta_{x_ix_j}=\frac{q_{x_ix_j}}{q_{x_i}}$. While {a} Markov process provides a straight forward framework for modeling the temporal behaviour of a random variable with finite states, it doesn't scale up well for large state spaces i.e. the size of intensity matrix, $\textbf{Q}_{x|\textbf{u}}$ grows exponentially with the number of variables. For example for a discrete random variable with $n=10$ states, it requires $2^{n=10}\simeq 1,024$ conditional intensities to be estimated. Thus to improve the scalability issue, the concept of conditional Markov process is introduced. 

\vspace{-1.0em}
\subsection{Conditional Markov Process}
\label{Subsection: conditional_markov_process}
To improve the scalability of Markov processes for large state spaces, Nodelman et al.\cite{nodelman_continuous_2002} introduced the idea of {the} conditional Markov process, in which the transition intensity matrix changes over time, but not as a direct function of time{,} rather as a function of the state values of some parent variable which also evolves as {a} Markov process. Let, $\textbf{u}=\{u_1,..,u_k\}$ represent the state space of the parent variable, then the conditional intensity matrix (CIM) $\textbf{Q}_{X|\textbf{u}}$ can be written as

\[
\textbf{Q}_{X|u} = \begin{bmatrix} 
    -q_{x_1|u}    & q_{x_1x_2|u}    & \dots & q_{x_1x_n|u} \\
    q_{x_2x_1|u}  & -q_{x_2|u}      & \dots & q_{x_2x_n|u} \\
    \vdots      & \vdots        & \ddots & \vdots\\
    q_{x_nx_1|u}  & q_{x_nx_2|u}    & \dots & -q_{x_n|u}
    \end{bmatrix}
\]

\noindent Conditioning the transitions on parent conditions increase the sparsity of the intensity matrix considerably, which is especially helpful for modeling large state spaces. When no parent variable is present, the CIM will be the same as the classic intensity matrix. When a parent variable $u$ is present, there will be an intensity matrix associated with each state of the parent variable $u\in \textbf{u}$. When multiple parent variables are present, there will be an intensity matrix associated with each combination of the states of the parent variables, which can still be represented by $\textbf{u}$.

For our case study in Section \ref{Section:CTBN_Case_Study}, we model the transition intensities between different states of multiple chronic conditions based on conditional Markov processes. 
The model formulates the probability of transition between the (discrete) states of the multiple chronic conditions as a continuous function of time, namely exponential distributing, with respect to the associate conditional intensities. Such formulation helps better to capture the actual progression of the multiple chronic conditions
(Please also see Figure \ref{Figure:CTBN_Learned_Structure} in Section \ref{Subsection:Learned_Bayesian_Structure} ). 

\vspace{-1.0em}
\subsection{Continuous Time Bayesian Network}
\label{Subsubsection:Continuous Time Bayesian Networks}
A continuous time Bayesian network (CTBN) is built by putting together a set of CIMs under a graph structure \cite{norris1998markov, nodelman_continuous_2002}. The two main components of a CTBN are: 
\begin{enumerate}
  \item An initial distribution $(P_x^0)$, which formulates the structure of the (conditional) relationship among the {random} variables and is specified as a Bayesian network.
  \item A state transition model $(Q_{X_i|\textbf{u}})$, which describes the transient behavior of each variable $x_i\in X$ given the state of parent variables $\textbf{u}$, and is specified based on CIMs.
\end{enumerate}
\noindent Each node $Y_i  (Y_i \in \textbf{X})$ in the CTBN is a random variable with finite discrete states $x_i=\{1,...,l\}$.
Each edge $x_i \to x_j$ in the graph implies the dynamics of variable $x_j$ evolution over time, which depends on the status of the parent variable $x_i$. As the graph suggests ($x_i \to x_j$), the $x_i$'s evolution cannot simultaneously depend on the status of the $x_j$ \cite{nodelman_continuous_2002}. CTBN explicitly represents the temporal dynamics of random variables, which enable the extraction of the probability distribution over time when a specific event occurs. Unlike traditional Bayesian networks, CTBN allows for cycles in graph, $\mathcal{G}$. This is an important property for modeling reinforcing loop between random variables, as we will show in the case study for modeling {the} relationship between multiple chronic conditions. Later, we will also use this property to develop a regularization based method for structure learning of the CTBN.

\vspace{-1.0em}
\subsection{Queries and Inference}
\label{Subsubsection:Inference_in_CTBN}
A CTBN is a homogeneous Markov {process} with a joint intensity matrix $\textbf{Q}_{x|\textbf{u}}$ that can be defined as
\begin{align}
\label{Equation:Probability density}
    \textbf{Q}_{x|\textbf{u}} = \prod_{X \in X} \textbf{Q}_{X|pa(X)}
\end{align}
Thus, any query can be answered by an explicit representation of a Markov process. Once the joint intensity matrix is formed, it can be used to answer queries the same way as a Markov process. Given a joint intensity matrix $Q_{x|v}$, the distribution $P_X^0$ over the state of $X$ at any time $t$ can be calculated using Equation \ref{Equation:Probability density2}.
\begin{align}
\label{Equation:Probability density2}
    P_x(t) &= P_x^0 exp(\textbf{Q}_{x|\textbf{u}}t)
\end{align}
To calculate the joint distribution over any two points in time, Equation \ref{Equation:Probability density2} can be modified as following-
\begin{align}
\label{Equation:Probability_density3}
    P_x(t,k) &= P_x^0 exp(\textbf{Q}_{x|\textbf{u}}(t-k)), \tab t \geq k
\end{align}
The inference operation can be performed using {either the} exact {or the} approximate algorithm. The use of amalgamation methods \cite{nodelman_continuous_2002} is an exact algorithm that involves large matrix representation. However, for systems with large state space, it becomes computationally inefficient (also, exact inference in CTBN is NP-Hard); thus, we tend to utilize the approximation methods \cite{nodelman2012expectation, shelton2010continuous}. Sampling-based algorithms can also be used to perform {the} inference operation. 
\vspace{-1.0em}
\subsection{Parameter Estimation}
\label{Subsubsection:Parameter_Estimation}
Having a dataset $\mathcal{D} = \{\tau_{h=1}, \tau_{h=2},....,\tau_{h=H}\}$ of $H$ observed transitions, where $\tau_h$ represents the time at which the $h^{th}$ transition has occurred, and $\mathcal{G}$ is a Bayesian network
defining the structure of the (conditional) relationship among variables, we can use maximum likelihood estimation (MLE) (Equation \ref{Equation:Likelihood_Function}) to estimate {the} parameters of the {CTBN} model as defined in Nodelman et al \cite{nodelman2002learning}

\begin{align}
\label{Equation:Likelihood_Function}
    \begin{split}
    L_x(\textbf{q}_{x|\textbf{u}}:\mathcal{D}) &= \prod_{\textbf{u}} \prod_x q_{x|\textbf{u}}^{M{[x|\textbf{u}]}} exp(-q_{x|\textbf{u}}T{[x|\textbf{u}]})
    \end{split}
\end{align}

\noindent where, $T[x|\textbf{u}]$ is the total time $X$ spends in the same state $x$, and $M{[x|\textbf{u}]}$ is the total number of times $X$ makes a transition out of state $x$ given, $x = x'$. The log-likelihood function can be then written as- 
\begin{align}
\label{Equation:Log-Likelihood_Function}
    \begin{split}
    l_x(q_{x|\textbf{u}}:\mathcal{D}) &= \sum_{\textbf{u}} \sum_x M[x|\textbf{u}]\hspace{1pt} ln(q_{x|\textbf{u}}) - q_{x|\textbf{u}}\hspace{1pt} T[x|\textbf{u}]
    \end{split}
\end{align}
Maximizing Equation \ref{Equation:Log-Likelihood_Function}, provides the maximum likelihood estimate of the conditional intensities as shown in Equation \ref{Equation:MLE}

\begin{align}
\label{Equation:MLE}
    \hat{q}_{x|\textbf{u}} = \frac{M[x|\textbf{u}]}{T[x|\textbf{u}]}
\end{align}
The above estimation is true for the case with complete data. For the cases including incomplete dataset, expectation maximization (EM) algorithms can be used \cite{nodelman2012expectation,nodelman2012expectation2}.
\vspace{-1.0em}
\section{Proposed Methodology}
\label{Section: Methodology}
We begin with formulating the conditional dependencies of the CTBN as {a} Poisson regression of some exogenous variables $\textbf{z}$. Next, we drive the likelihood function of the functional CTBN as a collection of Poisson regression likelihoods. {Afterwards}, we propose an adaptive group regularization method for structure and parameter learning of the functional CTBN. Finally, we present a post-processing an early stopping approach based on Gaussian mixture model clustering of the estimated parameters.

\vspace{-1.0em}
\subsection{Functional Continuous Time Bayesian Network with Conditional Dependencies as Poisson Regression}
\label{Subsubection:RegressionBasedAlgorithms}
In many real world problems, such as the progression of multiple chronic conditions, which is discussed in our case study, the evolution of the state variables (chronic conditions) not only depends on their immediate past state and the states of their parents variable (pre-existing conditions) but also (possibly) on some exogenous variables (socio-demographic factors). 

We propose to formulate the conditional intensities of the CTBN as a function of exogenous risk factors using a Poisson regression, which utilizes a special set of generalized linear models. Let $\textbf{z}=\{z_1,z_2,...,z_m\}$ denote a set of exogenous variables, i.e., patient-level risk factors such as age, gender, race, education, marital status, etc. The \textit{rate of transition} between any two-state variables, (say, chronic conditions) can be derived as:
\begin{equation}
\label{Equation:CTBN_Poisson_1}
	\begin{split}
            log(q_{x_{i} x_{j}|\textbf{u}}) &= \beta_{0_{x_{i} x_{j}|\textbf{u}}} + \beta_{1_{x_{i} x_{j}|\textbf{u}}}z_1 + ..... \\
            &\tab\tab {+}\: + \beta_{m_{x_i x_j|\textbf{u}}}z_m\\
                                                &=  \textbf{z}\boldsymbol{\beta}_{x_{i},x_{j}|\textbf{u}}
	\end{split}
\end{equation}
    
\begin{equation}
\label{Equation:CTBN_Poisson_2}
	\begin{split}
            log(q_{x_{i}|\textbf{u}}) &= \beta_{0_{x_i|\textbf{u}}} + \beta_{1_{x_i|\textbf{u}}}z_1 + ..... + \beta_{m_{x_i|\textbf{u}}}z_m \\
                                        &= \textbf{z}\boldsymbol{\beta}_{x_{i}|\textbf{u}}
	\end{split}
\end{equation}

\noindent where $\boldsymbol{\beta}_{x_{i}|\textbf{u}}=[\beta_{0_{x_i|\textbf{u}}} , ..., \beta_{m_{x_i|\textbf{u}}}]^\intercal$ and $\beta_{k_{x_i|\textbf{u}}} = \sum_{i \neq j} \beta_{k_{x_{i} x_{j}|\textbf{u}}}$ are the coefficients of the Poisson regression.
When the state space of the system and related conditions are binary (as in our case study on MCC transitions, where MCC states include having/not having each of the conditions), the conditional intensities in $\textbf{Q}_{x_i |\textbf{u}_i}$, can be estimated just using Equation  \ref{Equation:CTBN_Poisson_2} because for Markov processes with binary states $ q_{x_i |\textbf{u}} = - \sum_{j\neq i} q_{(x_i x_j |\textbf{u})} = -q_{(x_i x_j , j\neq i|\textbf{u})}$. This feature considerably simplifies the estimation of the functional CTBN conditional intensity matrix based on Poisson regression.
\vspace{-1.0em}
\subsection{Parameter Estimation}
\label{Subsubection:Parameter_Estimation_PRCTBN}
Having a dataset of state variables' transition trajectories, $\mathcal{D} = \{ \tau_{(p=1,h=1)} , ... , \tau_{(P,H)} \}$, where  $\tau_{(p,h)}$ represents the time at which the $h^{th}$ (MCC) transition of the $p^{th}$ subject has occurred, we can use maximum likelihood estimation (MLE) to estimate the parameters of the proposed functional CTBN. Assuming that all transitions are observed, the likelihood of $\mathcal{D}$ can be decomposed as the product of the likelihood for individual transitions, $q$. Let $d= \langle $\textbf{z}$, \textbf{u}, x_{i|\textbf{u}}, t_d,x_{j|\textbf{u}} \rangle$ {represents a state} transition {for} subject $p$ with risk factors $\textbf{z}$, and parent variables $u$, which/who made transition to state $x_{j|\textbf{u}}$ after spending the amount of time $t_d=\tau_{(p,h)}-\tau_{(p,h-1)}$ in state $x_{i|\textbf{u}}$. If the state space of {the} conditions is considered as binary, i.e. having/not having a chronic condition, the likelihood of the single transition {$d$} can be written as in Equation \ref{Equation:CTBN_Poisson_3}


\begin{align}
\label{Equation:CTBN_Poisson_3}
    \begin{split}
    L_x (\textbf{z}\boldsymbol{\beta}_{x_{i}|\textbf{u}}: \mathcal{D})&=\\
    &\prod_p \prod_h \prod_{\textbf{u}} \prod_{x_{i}} \textbf{z}\boldsymbol{\beta_{x_{i}|\textbf{u}}}\left[exp(-\textbf{z}\boldsymbol{\beta}_{x_{i}|\textbf{u}}   t_d[{x_{i|\textbf{u}}}])\right]
    \end{split}
\end{align}

By multiplying the likelihoods of {all} transitions for all subjects (patients) $(\tau_{(p,h)} \in \mathcal{D})$ and taking the \textit{log}, we obtain the overall log-likelihood function as in Equation \ref{Equation:CTBN_Poisson_4}

\begin{equation}
\label{Equation:CTBN_Poisson_4}
    \centering
    \begin{split}
        l_x (\boldsymbol{\beta}_{x_{i}|\textbf{u}}:\mathcal{D})    &=\\ &log [\prod_p \prod_h \prod_{\textbf{u}} \prod_{x_{i}} \textbf{z}\boldsymbol{\beta}_{x_{i}|\textbf{u}} exp(-\textbf{z}\boldsymbol{\beta}_{x_{i}|\textbf{u}}t_{d_p}^{h}[x_{i}|\textbf{u}]) ]\\
        &= \sum_p \sum_h \sum_{\textbf{u}} \sum_{x_i} (\textbf{z}_p 
        \boldsymbol{\beta}_{x_{i}|{\textbf{u}}}) - \\
        & \sum_p \sum_h \sum_{\textbf{u}} \sum_{x_{i}} \bigg\{ t_{d_p}^{h}[x_{i}|\textbf{u}] exp(\textbf{z}_p\boldsymbol{\beta}_{x_i|{\textbf{\textbf{u}}}}) \bigg\}
    \end{split}
\end{equation}

\begin{figure}[h!]
    \centering
    \includegraphics[scale = 0.15]{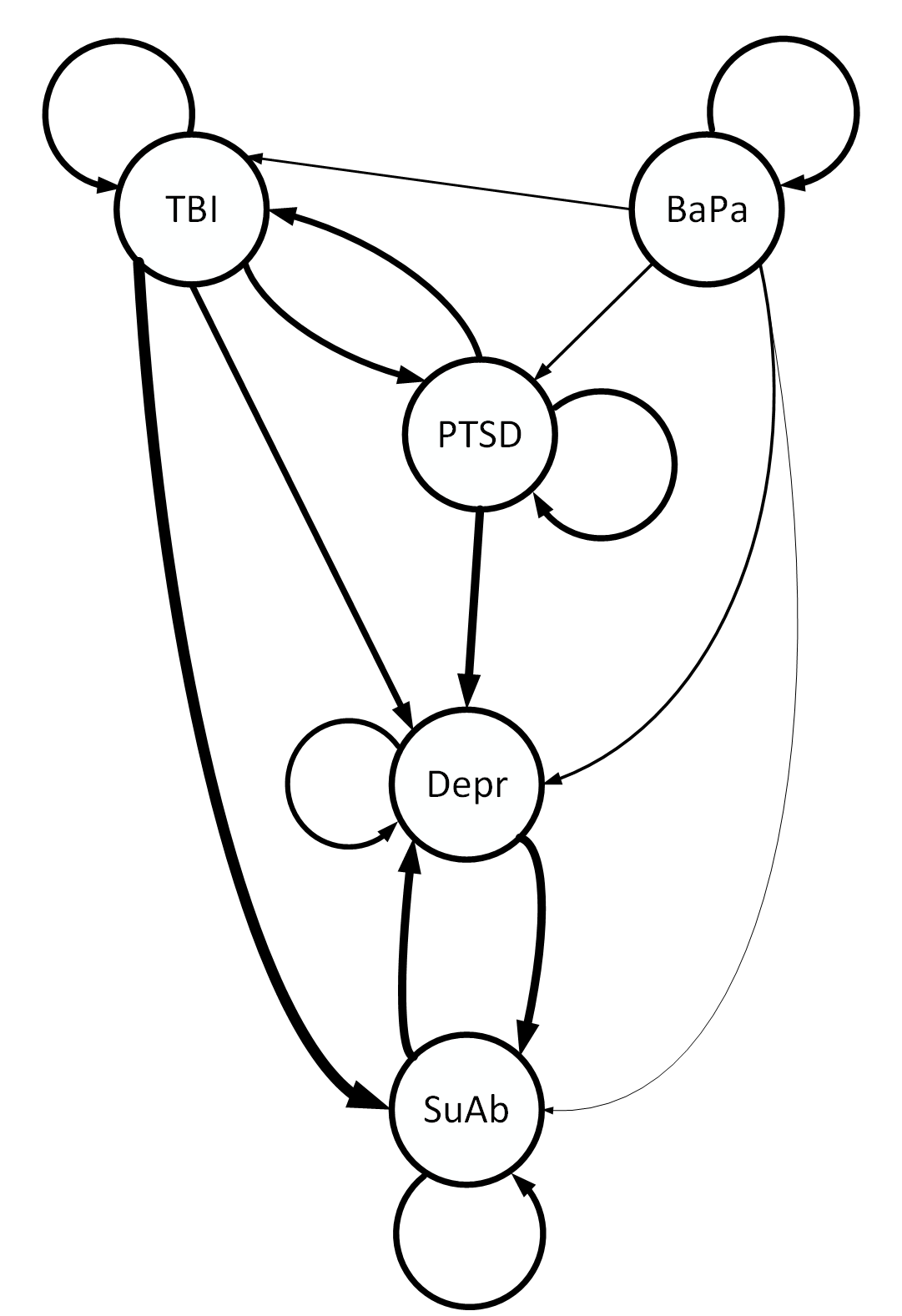}
    \caption{Illustration of  the functional CTBN for 5 MCC including Traumatic Brain Injury (TBI), Back Pain (BaPa), Post Traumatic Stress Disorder (PTSD), Depression (Depr), and Substance Abuse (SuAb)  based on {the case study}: The thickness of the edges represent the strength{s} of the conditional {intensities}, $q_{x|\textbf{u}}$.}
    \label{Figure:CTBN_Illustration_1}
\end{figure}

Equation \ref{Equation:CTBN_Poisson_4} is a convex function in terms of $\boldsymbol{\beta}_{x_{i}|{\textbf{u}}}$ and can be maximized using a convex optimization algorithm{s} such as Newton-Raphson (See Figure \ref{Figure:CTBN_Illustration_1}).

Given the structure of the functional CTBN, i.e. the parent set for each variable, the maximum number of parameters to be estimated in Equation \ref{Equation:CTBN_Poisson_4} will be $\Bar{\Bar{\textbf{x}}} \times \Bar{\Bar{\textbf{z}}} \times 2^{max(\Bar{\Bar{\textbf{u}}})+1}$, where $\Bar{\Bar{\textbf{x}}}$ is the number of state variables (conditions), $\Bar{\Bar{\textbf{z}}}$ is the number of exogenous variables (risk factors) presents in the system, and max($\Bar{\Bar{\textbf{u}}}$) is the maximum number of parents considered ({pre-existing} diseases for each condition). Therefore, as in classical Bayesian networks, the number of parents has a direct and exponential influence on the computational efficiency of the estimation process and should be limited to a small number. 

We propose to assume the conditional effect of parents is multiplicative, i.e. $q_{x_{i}|u_1,u_2}=q_{x_{i |u_1}}.q_{x_{i|u_2}}$, to make the conditional effect of the risk factors additive given the set of parents, i.e. $\boldsymbol{\beta}_{x_ix_j|\textbf{u}={u_1,...u_k}} = \boldsymbol{\beta}_{x_ix_j|u_1} + \ldots + \boldsymbol{\beta}_{x_ix_j|u_k}$. This assumption , which is on a par with the Noisy-OR \cite{pearl_1988_reasoning, Onisko_2001} and the CT-NOR \cite{simma_2008}, reduces the maximum number of parameters to be estimated to $\Bar{\Bar{\textbf{x}}} \times \Bar{\Bar{\textbf{z}}} \times 2 \times ({max(\Bar{\Bar{\textbf{u}}})+1})$). However, in situations where the number of exogenous variables (risk factors for the multiple chronic conditions) are large, the estimation of the Poisson regression parameters can still be computationally challenging, even with the multiplicative assumption. To address this problem, we propose to consider {principal} component analysis (PCA) or kernel PCA (KPCA) to first extract a few informative features of $\textbf{z}$, and then use those features of the original covariates for building the Poisson regression model for each conditional {intensity} \cite{alaeddini2019integrated}. While reducing the interpretability of the estimated parameters ($\boldsymbol{\beta}_{x|\textbf{u}}$), using dimensionality reduction (PCA or KPCA) helps with {efficient modeling of} the non/linear relationship among the risk factors. Considering the PCA for our case study, while it is restricted to only linear correlation between (exogenous) variables, Our analysis shows the PCA captures a considerable portion of variation (\textgreater 82\%) in our dataset using only the first principal component, which is also evidenced in the proposed methodology performance as described in Section \ref{Section:CTBN_Case_Study}.

\vspace{-1.0em}
\subsection{Adaptive Group Regularization Framework for Structure Learning in Continuous Time Bayesian Network}
\label{Subsubsection:AdaptiveGroupRegularization}
The parameter estimation approach presented above requires the parent set of each condition to be known, which is equivalent to knowing the structure of the Bayesian network. 
Here, we propose an adaptive group regularization-based framework to simultaneously learn the structure ($\mathcal{G}$) and conditional intensities $(\textbf{Q}_{x_i|\textbf{u}})$ of the functional Bayesian network model. Regularization-based structure learning is a recent approach that is gaining popularity for parameter estimation in graphical models\cite{lee2007efficient, wainwright2007high, schmidt2007learning}. However, since regularization can result in cycles in graphical models, it is not generally considered for directed graphs. Given that the proposed functional CTBN has a special structure based on a conditional intensity matrix that allows for cycles, this study proposes to extend the regularization-based structure learning to functional CTBN.

\begin{figure}[!h]
    \centering
    \includegraphics[scale = 0.125]{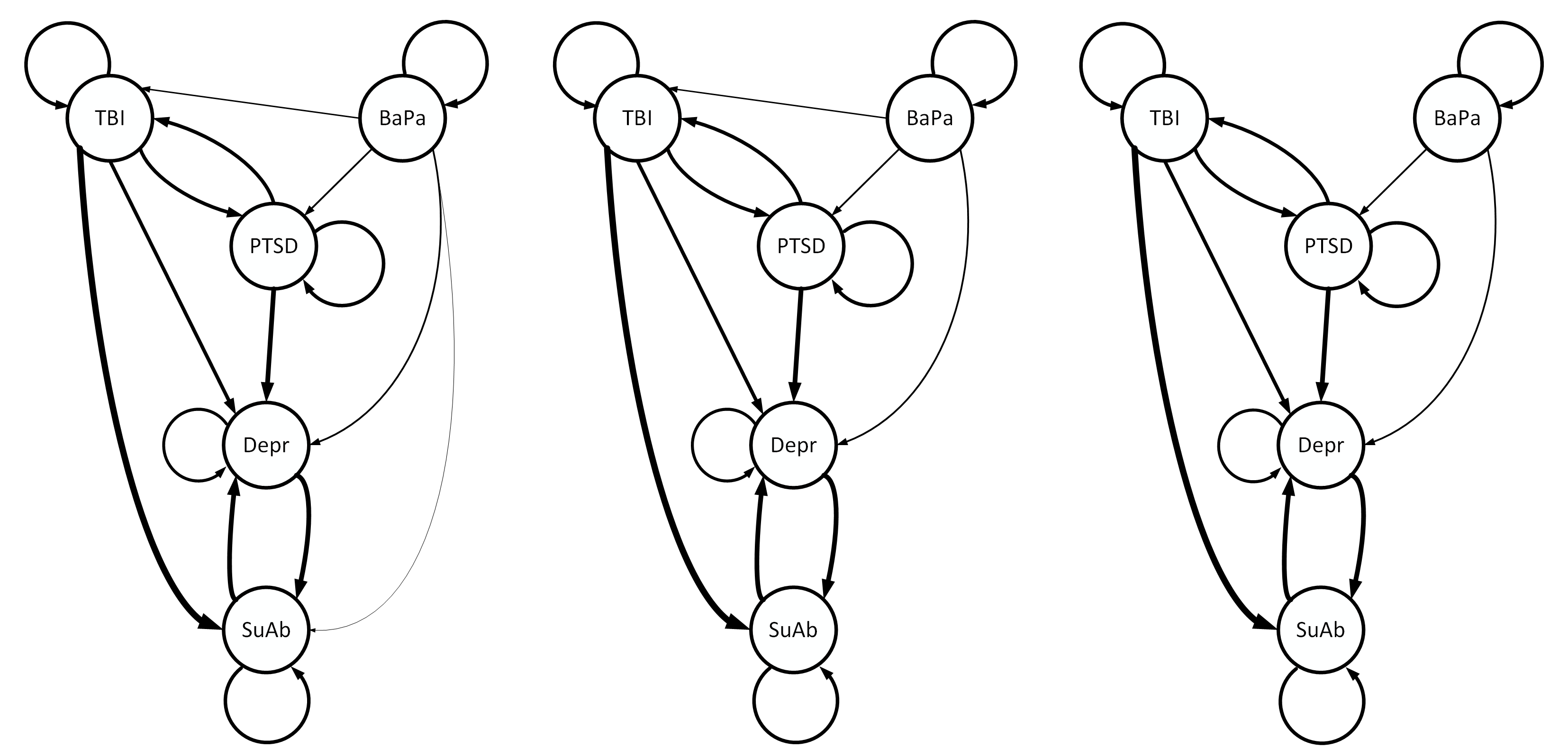}
    \caption{
    The effect of changing the tuning parameter (regularization path) on the structure of the {functional} CTBN for {the 5 MCC in our case study including TBI, BaPA, PTSD, Depr, and SuAb.}
    }
    \label{Figure:CTBN_Illustration_2}
    \vspace{-1em}
\end{figure}

\begin{figure}[h!]
    \centering
    \includegraphics[scale = 0.30]{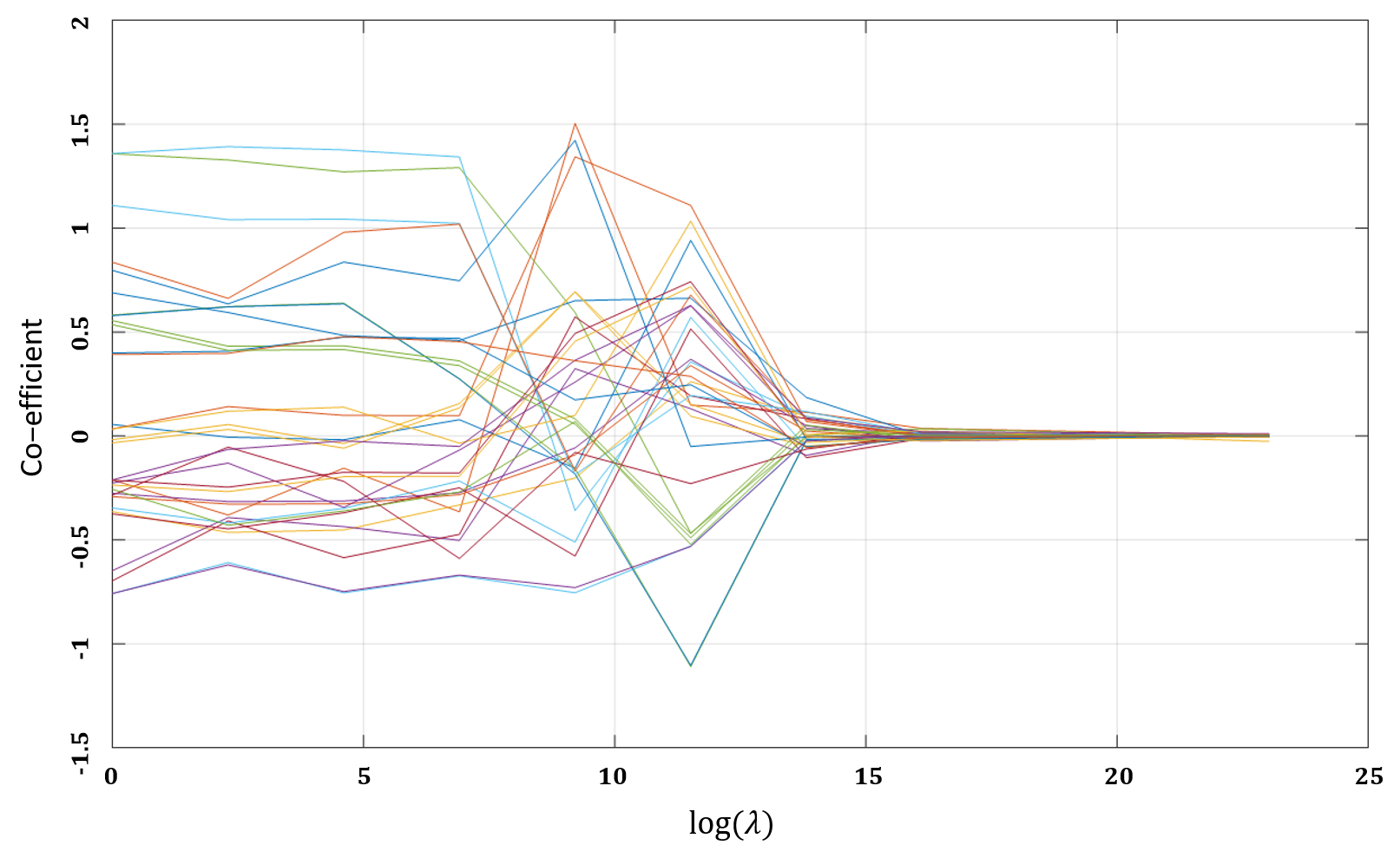}
    \caption{ {The regularization path of the tuning parameter for the proposed model (For the sake of simplicity only some of the total learned parameters are shown).}}
    \label{Figure:Regularization_path}
\end{figure}

Considering the negative log likelihood of the fully connected functional CTBN, we propose to add an adaptive group regularization term to the negative log likelihood function to penalize groups of parameters pertaining to each specific conditional intensities as in Equation \ref{Equation:CTBN_Minimization_1}.

\begin{equation}
    \label{Equation:CTBN_Minimization_1}
    \centering
    \minA -l_x(\textbf{q}:\mathcal{D}) + k \sum_{x_{i}|\textbf{u}}\lambda_j\|\boldsymbol{\beta}_{x_{i}|\textbf{u}}\|
\end{equation}

\noindent where, $\|\boldsymbol{\beta}_{x_{i}|\textbf{u}}\| = \sqrt{\sum_{\textbf{u}} \sum_{x_{i}} (\boldsymbol{\beta}_{x_{i|\textbf{u}}}.\boldsymbol{\beta}^T_{x_{i|\textbf{u}}} )}$ is the norm of the group of parameters associated with each conditional intensity, $k$ is the groups size which is based on the number of coefficients in the Poisson regression for each conditional intensity{,} $\lambda_j = \lambda \|\Tilde{\boldsymbol{\beta}_j}\|^{-1}$ is the tuning parameters of the adaptive group regularization that control the amount of shrinkage, where $\lambda$ is inversely weighted based on the unpenalized estimated value of the regression coefficients $\Tilde{\boldsymbol{\beta}_j}$  \cite{Wang2008ANO}. The index $j$ implies the adaptive penalization applied to each grouped parameter.
Fast-iterative shrinkage thresholding algorithm (FISTA) can be used for solving Equations \ref{Equation:CTBN_Minimization_1} \cite{alaeddini2017multi}. 
Figure \ref{Figure:Regularization_path} shows the regularization path of the tuning parameter for some of the parameters of the proposed model.

An interesting feature of the adaptive group-regularization based structure learning is that we can use the regularization path to control the level of sparsity in the proposed functional CTBN (See Figure \ref{Figure:CTBN_Illustration_2}).

\vspace{-1.2em}
\subsection{Post Processing and Early Stopping}
\label{Subsebsection:Early_Stopping_Criteria}
Each of the conditional dependencies (edges) {in} the proposed functional CTBN is consisted of {a} Poisson regressions with multiple exogenous variables.
For cases where the number of parent variables and/or the exogenous variables are large, the process of structure and parameter learning requires substantial computation. However, the majority of changes in the estimated values of the (Poisson regression) parameters happen in the early iterations of the learning algorithm. The (numerous) remaining iterations of the learning algorithms make minor adjustments to the estimated values of the parameters, especially pushing parameters with a small value toward zero. These later steps to push insignificant parameters toward zero can take many iterations without significantly changing other (significant) parameter values.  

Meanwhile, for a sufficiently large choice of the tuning parameter, some of the parameters will be zero. This is because for a general regression problem, setting the tuning parameter to infinity results in all coefficients except the intercept to be zero. On the other hand, a very small choice of the tuning parameter can result in all estimated parameters being non-zero. Additionally, having all the parameters in the proposed model as non-zero would be equivalent to having a fully connected functional CTBN network, which is not expected in most cases. Therefore, it is plausible to have some zero parameters to be discovered by an appropriate regularization method, such as the adaptive group regularization framework as described in section \ref{Subsubsection:AdaptiveGroupRegularization}.

To reduce the number of (additional) iterations for zeroing non-significant parameters, we propose to use Gaussian mixture models (GMM).
For this purpose, we stop the learning algorithm when there is {no} significant change in the estimated parameters. Next, we use a GMM to model the estimated parameters \cite{mclachlan2000finite}. The GMM is expected to have one cluster with a mean around zero (representing the {insignificant} parameters to be pushed toward zero) and one or more clusters with non-zero means (representing significant parameters). 
Once the clusters and their parameters are identified, we choose the cluster with (around) zero mean and assign a value of zero to all parameters within $\pm 3\sigma$ (standard deviation) around the mean $\mu\sim 0$.
We run an additional iteration of the learning algorithm with the zero\'d parameters to ensure convergence.
\begin{table*}[hb!]
    \centering
  	\caption{Demographics of the patients included in the study.}
  	\begin{adjustwidth}{0in}{0in}
    	\scalebox{0.65}{
        	\begin{tabular}{|c|c|c|c|c|c|c|c|c|c|c|c|c|c|c|c|}
    
    \hline
    \multicolumn{1}{|l|}{Sl No.} & \multicolumn{1}{|l|}{Race} & \multicolumn{2}{|c|}{Gender} & \multicolumn{2}{|c|}{Marital Status} & \multicolumn{4}{|c|}{Age Group} & \multicolumn{6}{|c|}{Education} \\
    \hline
          &       & Male  & Female & Married  & Un-Married & 18-30 & 31-40 & 41-50 & 51- Rest & Unknown & $<$ High School & High School & Some College & College Graduate & Post College\\
    
    \hline     
    1     & \multicolumn{1}{l|}{White} & 148355 & 19183 & 74487 & 93051 & 96799 & 36003 & 26167 & 8569  & 2334  & 2037  & 129921 & 16743 & 12024 & 4479 \\
          &       & \cellcolor[rgb]{ .851,  .851,  .851} 57.58\% & \cellcolor[rgb]{ .851,  .851,  .851} 7.45\% & \cellcolor[rgb]{ .851,  .851,  .851} 28.91\% & \cellcolor[rgb]{ .851,  .851,  .851} 36.12\% & \cellcolor[rgb]{ .851,  .851,  .851} 37.57\% & \cellcolor[rgb]{ .851,  .851,  .851} 13.97\% & \cellcolor[rgb]{ .851,  .851,  .851} 10.16\% & \cellcolor[rgb]{ .851,  .851,  .851} 3.33\% & \cellcolor[rgb]{ .851,  .851,  .851} 0.91\% & \cellcolor[rgb]{ .851,  .851,  .851} 0.79\% & \cellcolor[rgb]{ .851,  .851,  .851} 50.43\% & \cellcolor[rgb]{ .851,  .851,  .851} 6.50\% & \cellcolor[rgb]{ .851,  .851,  .851} 4.67\% & \cellcolor[rgb]{ .851,  .851,  .851} 1.74\% \\
          
    \hline
    2     & \multicolumn{1}{l|}{Black} & 35758 & 11828 & 23308 & 24278 & 20047 & 12468 & 12710 & 2361  & 658   & 504   & 37506 & 4819  & 3160  & 939 \\
          &       & \cellcolor[rgb]{ .851,  .851,  .851} 13.88\% & \cellcolor[rgb]{ .851,  .851,  .851} 4.59\% & \cellcolor[rgb]{ .851,  .851,  .851} 9.05\% & \cellcolor[rgb]{ .851,  .851,  .851} 9.42\% & \cellcolor[rgb]{ .851,  .851,  .851} 7.78\% & \cellcolor[rgb]{ .851,  .851,  .851} 4.84\% & \cellcolor[rgb]{ .851,  .851,  .851} 4.93\% & \cellcolor[rgb]{ .851,  .851,  .851} 0.92\% & \cellcolor[rgb]{ .851,  .851,  .851} 0.26\% & \cellcolor[rgb]{ .851,  .851,  .851} 0.20\% & \cellcolor[rgb]{ .851,  .851,  .851} 14.56\% & \cellcolor[rgb]{ .851,  .851,  .851} 1.87\% & \cellcolor[rgb]{ .851,  .851,  .851} 1.23\% & \cellcolor[rgb]{ .851,  .851,  .851} 0.36\% \\
          
    \hline
    3     & \multicolumn{1}{l|}{Hispanic} & 25373 & 4232  & 14523 & 15082 & 17016 & 6606  & 4758  & 1225  & 386   & 360   & 23592 & 2933  & 1893  & 441 \\
          &       & \cellcolor[rgb]{ .851,  .851,  .851} 9.85\% & \cellcolor[rgb]{ .851,  .851,  .851} 1.64\% & \cellcolor[rgb]{ .851,  .851,  .851} 5.64\% & \cellcolor[rgb]{ .851,  .851,  .851} 5.85\% & \cellcolor[rgb]{ .851,  .851,  .851} 6.60\% & \cellcolor[rgb]{ .851,  .851,  .851} 2.56\% & \cellcolor[rgb]{ .851,  .851,  .851} 1.85\% & \cellcolor[rgb]{ .851,  .851,  .851} 0.48\% & \cellcolor[rgb]{ .851,  .851,  .851} 0.15\% & \cellcolor[rgb]{ .851,  .851,  .851} 0.14\% & \cellcolor[rgb]{ .851,  .851,  .851} 9.16\% & \cellcolor[rgb]{ .851,  .851,  .851} 1.14\% & \cellcolor[rgb]{ .851,  .851,  .851} 0.73\% & \cellcolor[rgb]{ .851,  .851,  .851} 0.17\% \\
          
    \hline
    4     & \multicolumn{1}{l|}{Asian} & 5639  & 981   & 3067  & 3553  & 3235  & 1361  & 1564  & 460   & 131   & 60    & 4732  & 598   & 879   & 220 \\
          &       & \cellcolor[rgb]{ .851,  .851,  .851} 2.19\% & \cellcolor[rgb]{ .851,  .851,  .851} 0.38\% & \cellcolor[rgb]{ .851,  .851,  .851} 1.19\% & \cellcolor[rgb]{ .851,  .851,  .851} 1.38\% & \cellcolor[rgb]{ .851,  .851,  .851} 1.26\% & \cellcolor[rgb]{ .851,  .851,  .851} 0.53\% & \cellcolor[rgb]{ .851,  .851,  .851} 0.61\% & \cellcolor[rgb]{ .851,  .851,  .851} 0.18\% & \cellcolor[rgb]{ .851,  .851,  .851} 0.05\% & \cellcolor[rgb]{ .851,  .851,  .851} 0.02\% & \cellcolor[rgb]{ .851,  .851,  .851} 1.84\% & \cellcolor[rgb]{ .851,  .851,  .851} 0.23\% & \cellcolor[rgb]{ .851,  .851,  .851} 0.34\% & \cellcolor[rgb]{ .851,  .851,  .851} 0.09\% \\
          
    \hline
    5     & \multicolumn{1}{l|}{Native} & 3081  & 707   & 1747  & 2041  & 2115  & 925   & 564   & 184   & 60    & 60    & 3004  & 376   & 217   & 71 \\
          &       & \cellcolor[rgb]{ .851,  .851,  .851} 1.20\% & \cellcolor[rgb]{ .851,  .851,  .851} 0.27\% & \cellcolor[rgb]{ .851,  .851,  .851} 0.68\% & \cellcolor[rgb]{ .851,  .851,  .851} 0.79\% & \cellcolor[rgb]{ .851,  .851,  .851} 0.82\% & \cellcolor[rgb]{ .851,  .851,  .851} 0.36\% & \cellcolor[rgb]{ .851,  .851,  .851} 0.22\% & \cellcolor[rgb]{ .851,  .851,  .851} 0.07\% & \cellcolor[rgb]{ .851,  .851,  .851} 0.02\% & \cellcolor[rgb]{ .851,  .851,  .851} 0.02\% & \cellcolor[rgb]{ .851,  .851,  .851} 1.17\% & \cellcolor[rgb]{ .851,  .851,  .851} 0.15\% & \cellcolor[rgb]{ .851,  .851,  .851} 0.08\% & \cellcolor[rgb]{ .851,  .851,  .851} 0.03\% \\
          
    \hline
    6     & \multicolumn{1}{l|}{Unknown} & 2135  & 361   & 1346  & 1150  & 1062  & 625   & 673   & 136   & 51    & 22    & 1808  & 287   & 223   & 105 \\
          &       & \cellcolor[rgb]{ .851,  .851,  .851} 0.83\% & \cellcolor[rgb]{ .851,  .851,  .851} 0.14\% & \cellcolor[rgb]{ .851,  .851,  .851} 0.52\% & \cellcolor[rgb]{ .851,  .851,  .851} 0.45\% & \cellcolor[rgb]{ .851,  .851,  .851} 0.41\% & \cellcolor[rgb]{ .851,  .851,  .851} 0.24\% & \cellcolor[rgb]{ .851,  .851,  .851} 0.26\% & \cellcolor[rgb]{ .851,  .851,  .851} 0.05\% & \cellcolor[rgb]{ .851,  .851,  .851} 0.02\% & \cellcolor[rgb]{ .851,  .851,  .851} 0.01\% & \cellcolor[rgb]{ .851,  .851,  .851} 0.70\% & \cellcolor[rgb]{ .851,  .851,  .851} 0.11\% & \cellcolor[rgb]{ .851,  .851,  .851} 0.09\% & \cellcolor[rgb]{ .851,  .851,  .851} 0.04\% \\
    
    \hline
          
			\end{tabular}%
		}
  	\label{Table:Demographics}%
    \end{adjustwidth}
\end{table*}%

\vspace{-1.0em}
\section{Case Study: Identifying Patterns of Multiple Chronic Conditions}
\label{Section:CTBN_Case_Study}
Long-lasting diseases, otherwise known as chronic conditions, can be considered a staple example of degradation processes that progress over time and contribute to the development of other new chronic conditions. The presence of two or more chronic medical conditions in an individual is commonly defined as multimorbidity, or multiple chronic conditions (MCC) \cite{wallace_managing_2015}. Here, we use the proposed functional CTBN to find the impact of patient level risk factors on the conditional dependencies of MCC and the evolution of different chronic conditions over time.

\vspace{-1.0em}
\subsection{Study Population and Demographics}
\label{Subsection:Study_Population}
The proposed case study dataset includes 608,503 patients with two or more MCC (including Traumatic Brain Injury (TBI), Post Traumatic Stress Disorder (PTSD), Depression (Depr), Substance Abuse (SuAb), and Back pain (BaPa)) who received medical care from the Department of Veteran Affairs for at least three years between 2002-2015. For meaningful prediction, we have removed the data for patients whose data was not maintained over three years. The dropout of patient information may be caused by but not limited to death, not requiring care or receiving care, etc. After dropping such data, the number of patients considered for the analysis is 257,633. The dataset includes the ICD-9-CM diagnosis codes documented during the course of VA care, during each inpatient or outpatient encounter. The risk factors (exogenous variables) considered in the dataset include age at VA entry, sex, race/ethnicity (White, African American, Hispanic, Asian/Pacific Islander, Native American, unknown), and education (less than high school, high school graduate, some college, college graduate, post-college education). Table \ref{Table:Demographics} shows the summary of the collected data based on patients' demographics. In this study, in order to reduce the computational complexity of the algorithm and to show the application of the dimensionality reduction technique, we use PCA to reduce the number of risk factors into one.

\vspace{-1.0em}
\subsection{Diagnosed Health Conditions}
\label{Subsection:Diagnosed_Health_Conditions}
We used ICD-9-CM codes from the inpatient and outpatient data (excluding ancillary and telephone care) to identify Traumatic brain injury (TBI), Post Traumatic Disorder (PTSD), Depression (Depr), substance abuse (SuAb), Back pain (BaPa) using validated published algorithms \cite{selim2004comorbidity}. PTSD, SuAb, and BaPa required two diagnoses at least seven days apart, while TBI, which is an acute injury, required only a single diagnosis. Each condition was coded as “0” or “1” for each year of care, with 1 indicating a diagnosis for that condition regardless of the number of instances for which each condition was diagnosed (Additional information on ICD-9 codes for the considered conditions can be found on appendix B). 

\begin{figure*}[t!]
    \vspace{-1.5em}
	\begin{center}
		\subfloat[Selection of {the optimum value for the} tuning parameter ($\lambda$) based on cross validation criteria.]{\label{Figure:Lambda_Tuning} 
		\includegraphics[width=.46\textwidth]{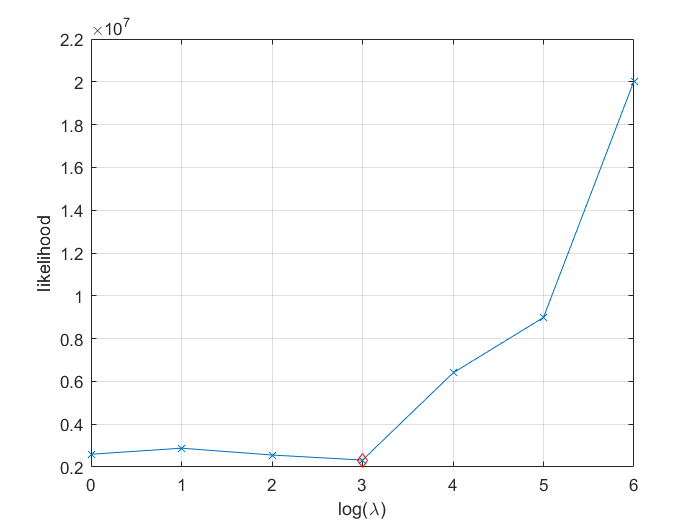}}
		\qquad
		\subfloat[{A GMM model fitted to} the estimated parameters of the functional CTBN; The red line identifies the mean, $\mu \sim 0$, and the green lines identify the $\pm 3\sigma$ bounds around the mean of the cluster with {$\mu \sim 0$}.]{\label{Figure:GMM_1} \includegraphics[width=.47\textwidth]{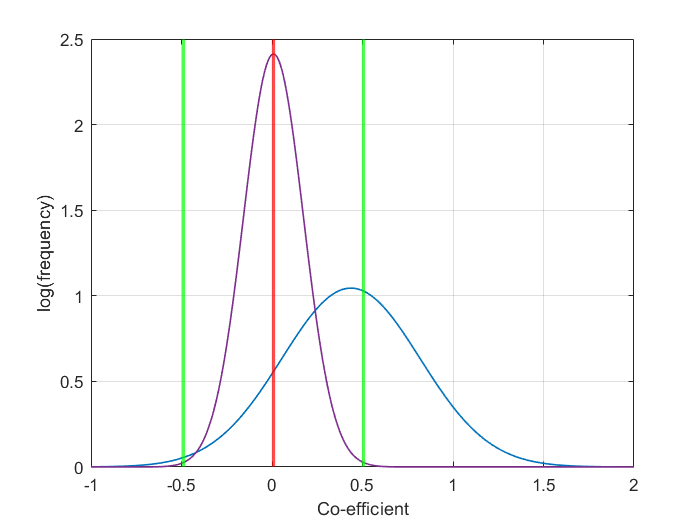}}
		\qquad
		\subfloat[The heatmap of the estimated parameters for the learned {functional CTBN} model. The {left hand side} matrix contains {all possible combinations} of parents and child {nodes interactions}. For example, the first {set of rows} (first 32 rows) of the matrix represents the parameters learned for {child node 1 and parents nodes 2, 3, 4, and 5}. The right hand side figure shows all possible {combinations of the parents and child nodes and the associate estimated parameters (1 for the presence of a condition and 0 for the absence of the condition)}.]{\label{Figure:Transition_Explain} \includegraphics[width=1\textwidth]{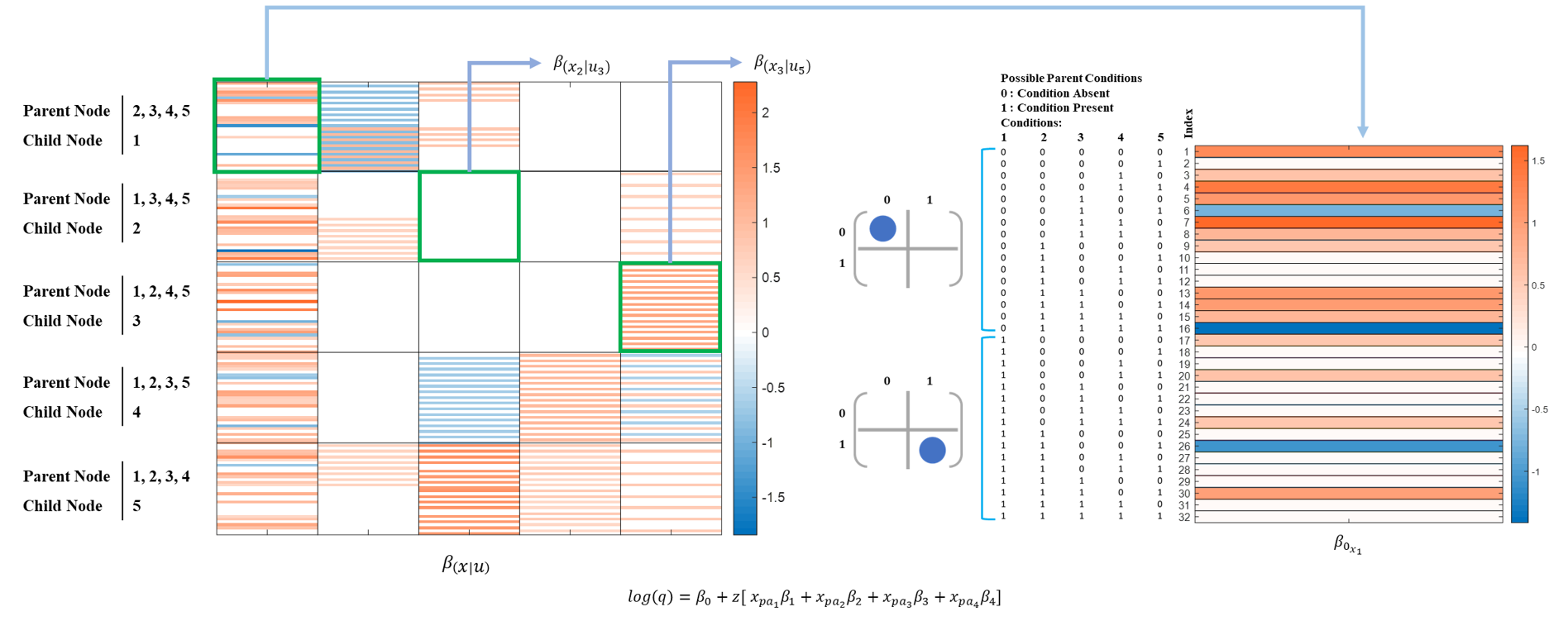}}
		\end{center}
	\caption{(a) Tuning of the hyper parameter ($\lambda$) based on cross validation, (b) Post processing and early stopping of the structure and parameter learning process using Gaussian mixture model, and (c) The estimated parameters of functional CTBN based on the optimal value of {the tuning parameter}.}
	\vspace{-1.0em}
\end{figure*}

\vspace{-1.0em}
\subsection{Structure and Parameter Learning }
\label{Subsection:Learned_Bayesian_Structure}
To identify the optimal value of the tuning parameter ($\lambda$) of the group regularization method for structure and parameter learning, we use cross-validation error based on several $\lambda$ values ($0,10^0, 10^1, 10^2,....,10^6$). Figure \ref{Figure:Lambda_Tuning} shows the cross-validation error for different $\lambda$ values.

We attain the structure of the functional CTBN and the conditional intensities based on the parameters estimated using the optimal value of $\lambda = 10^3$. Figure \ref{Figure:Transition_Explain} illustrates the heatmap of the estimated parameters ($\beta$) of the proposed CTBN based on $\lambda = 10^3$. As shown in figure \ref{Figure:Lambda_Tuning} and \ref{Figure:Transition_Explain}, setting $\lambda = 10^3$ not only provides considerably low (cross validation) error, but also significantly reduces the number of (non zero) parameters ({a} sparsity ratio of 64.75\%).

Figure \ref{Figure:Transition_Explain} {provides} the heatmap of the estimated parameters of the learned functional CTBN model, which is equivalent to the graphical model presented in figure \ref{Figure:CTBN_Illustration_1}. To identify the final structure of the functional CTBN, considering the sparse learned parameter matrix in figure \ref{Figure:Transition_Explain}, if all parameters (coefficients) of the Poisson regression connecting a parent node to a child node are zero, there exist{s} no edge between them. On the other hand, if there exists a non zero parameter for the Poisson regression {model} connecting a parent node to a child node, there exist{s} an edge between the two nodes, where the strength of the connection {is} represented by the conditional intensity value. 

Meanwhile, to reduce the number of training iterations for obtaining the spare matrix in Figure \ref{Figure:Transition_Explain}, we use GMM as explained in section \ref{Subsebsection:Early_Stopping_Criteria}.
Figure \ref{Figure:GMM_1} shows the Gaussian densities fitted to the estimated parameters at iteration 30,000 of the learning algorithm, which shows two clusters including one with zero mean and small variation (non significant parameters), and the other with non zero mean and high variance ({significant parameters}). We assign a value of zero to all parameters within $\pm 3\sigma$ (standard deviation) of the cluster with {the mean around zero}. We have verified this result by running the learning algorithm for an additional 20,000 iterations.
\begin{figure*}[t!]
    \centering
    \includegraphics[scale = 0.18]{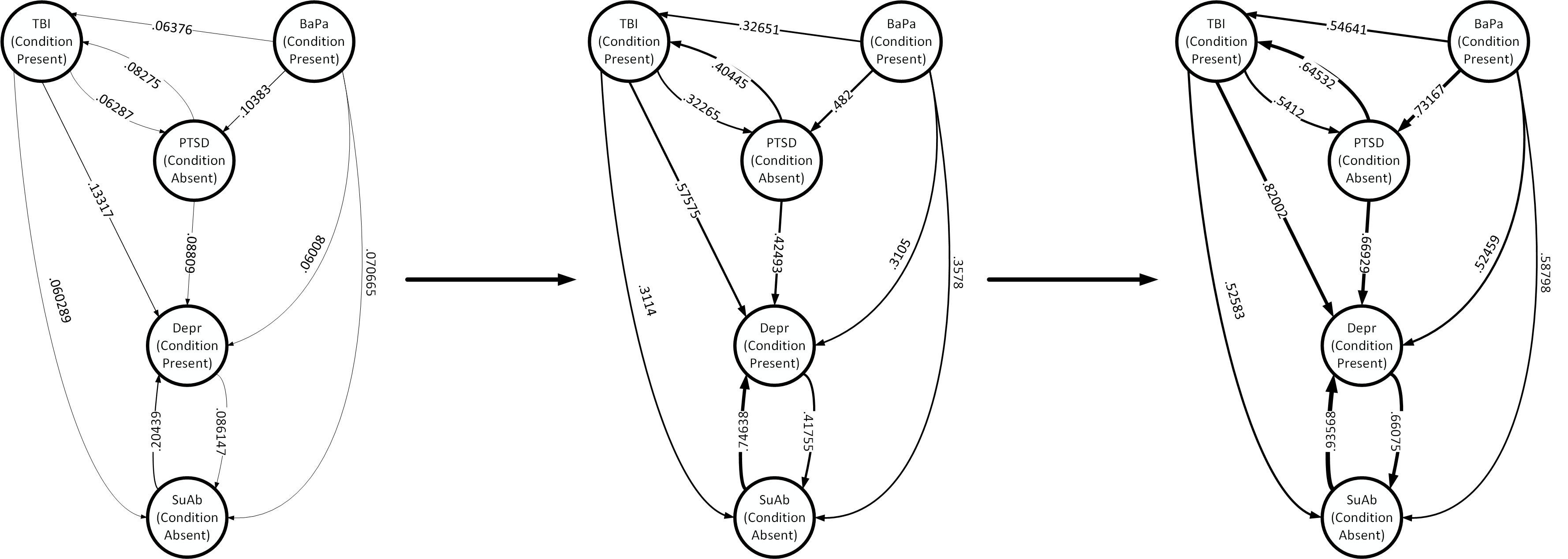}
    \caption{Learned functional CTBN structure for a set of given condition and their progression over time.}
    \label{Figure:CTBN_Learned_Structure}
    \vspace{-.5em}
\end{figure*}

Additionally, the functional CTBN allows for loop in the structure (as shown in figure \ref{Figure:CTBN_Illustration_1}). This is an important feature in studying MCC, because an MCC condition can simultaneously be the cause and/or the effect (result) of another MCC condition, i.e., depression and substance abuse. The functional CTBN also allows for the self-loops to represent staying in the same MCC state (existing/parent conditions) over (fixed amount of) time \cite{alaeddini_mining_2017}. 

Figure \ref{Figure:CTBN_Learned_Structure} illustrates the dynamics of the transition probabilities of MCC, namely the risk of acquiring a new condition over a period of three years for a sample patient with preexisting conditions TBI, BaPa, and Depression. As shown in the figure, the transition probabilities change as a function of time, which is intuitive in the presence of the preexisting conditions TBI, BaPa, and Depression. In particular, having the preexisting conditions TBI, BaPa, and Depression increases the likelihood of acquiring the new disorders PTSD and SuAb over time. It also increases the reinforcing loop between the existing conditions, which is also intuitive. This can help health practitioners and patients to better the short- and long-term (negative) impact of MCC on acquiring new conditions.

\begin{table*}[!b]
  \centering
  \caption{The AUC performance (of ROC) of the Functional CTBN (FCTBN) model for predicting the future in comparision to MTBN and LRMCL}
  \begin{adjustwidth}{0in}{0in}
  \scalebox{0.78}{
    \begin{tabular}{|c|c|c|c|c|c|c|c|c|c|c|c|c|c|c|c|}
    \hline
          & \multicolumn{3}{c|}{\textbf{Depression}} & \multicolumn{3}{c|}{\textbf{Substance Abuse}} & \multicolumn{3}{c|}{\textbf{PTSD}} & \multicolumn{3}{c|}{\textbf{Back Pain}} & \multicolumn{3}{c|}{\textbf{TBI}} \\
    \hline
          & \textbf{FCTBN} & \textbf{MTBN} & \textbf{LRMCL} & \textbf{FCTBN} & \textbf{MTBN} & \textbf{LRMCL} & \textbf{FCTBN} & \textbf{MTBN} & \textbf{LRMCL} & \textbf{FCTBN} & \textbf{MTBN} & \textbf{LRMCL} & \textbf{FCTBN} & \textbf{MTBN} & \textbf{LRMCL} \\
    \hline
    \textbf{Year 2} & \cellcolor[rgb]{ 1,  .6,  0} 75.89\% & 66.92\% & \cellcolor[rgb]{ 1,  .922,  .518} 67.34\% & \cellcolor[rgb]{ 1,  .6,  0} 76.61\% & \cellcolor[rgb]{ 1,  .922,  .518} 72.09\% & 71.91\% & \cellcolor[rgb]{ 1,  .6,  0} 79.72\% & \cellcolor[rgb]{ 1,  .922,  .518} 78.31\% & 67.02\% & \cellcolor[rgb]{ 1,  .6,  0} 73.53\% & 64.28\% & \cellcolor[rgb]{ 1,  .922,  .518} 66.35\% & 65.72\% & \cellcolor[rgb]{ 1,  .6,  0} 72.11\% & \cellcolor[rgb]{ 1,  .922,  .518} 67.59\% \\
    \hline
    \textbf{Year 3} & \cellcolor[rgb]{ 1,  .6,  0} 70.98\% & \cellcolor[rgb]{ 1,  .922,  .518} 65.96\% & 56.12\% & \cellcolor[rgb]{ 1,  .6,  0} 72.34\% & \cellcolor[rgb]{ 1,  .922,  .518} 69.09\% & 59.54\% & \cellcolor[rgb]{ 1,  .6,  0} 75.75\% & \cellcolor[rgb]{ 1,  .922,  .518} 74.95\% & 63.88\% & \cellcolor[rgb]{ 1,  .6,  0} 69.65\% & \cellcolor[rgb]{ 1,  .922,  .518} 62.16\% & 56.48\% & \cellcolor[rgb]{ 1,  .922,  .518} 63.36\% & \cellcolor[rgb]{ 1,  .6,  0} 70.08\% & 57.03\% \\
    \hline
    \textbf{Year 4} & \cellcolor[rgb]{ 1,  .6,  0} 68.92\% & \cellcolor[rgb]{ 1,  .922,  .518} 64.70\% & 49.61\% & \cellcolor[rgb]{ 1,  .6,  0} 70.75\% & \cellcolor[rgb]{ 1,  .922,  .518} 68.02\% & 53.00\% & \cellcolor[rgb]{ 1,  .6,  0} 73.76\% & \cellcolor[rgb]{ 1,  .922,  .518} 73.36\% & 62.25\% & \cellcolor[rgb]{ 1,  .6,  0} 67.79\% & \cellcolor[rgb]{ 1,  .922,  .518} 55.32\% & 50.18\% & \cellcolor[rgb]{ 1,  .922,  .518} 63.33\% & \cellcolor[rgb]{ 1,  .6,  0} 69.19\% & 51.02\% \\
    \hline
    \textbf{Year 5} & \cellcolor[rgb]{ 1,  .6,  0} 67.40\% & \cellcolor[rgb]{ 1,  .922,  .518} 64.11\% & 46.44\% & \cellcolor[rgb]{ 1,  .6,  0} 69.54\% & \cellcolor[rgb]{ 1,  .922,  .518} 59.41\% & 42.87\% & \cellcolor[rgb]{ 1,  .6,  0} 72.38\% & \cellcolor[rgb]{ 1,  .922,  .518} 71.92\% & 61.18\% & \cellcolor[rgb]{ 1,  .6,  0} 66.57\% & \cellcolor[rgb]{ 1,  .922,  .518} 62.66\% & 48.08\% & \cellcolor[rgb]{ 1,  .922,  .518} 62.92\% & \cellcolor[rgb]{ 1,  .6,  0} 69.56\% & 49.21\% \\
    \hline
    \end{tabular}%
    }
  \end{adjustwidth}
  \label{Table:CTBN_ROC}%
  \vspace{-1em}
\end{table*}%
\vspace{-1.0em}
\subsection{Performance Evaluation: Predictive Power}
\label{Performance_Comparison}
We utilize the validation set method based on 250\,000 patients for training and 7\,633 patients for validation, along with the Area Under the Curve (AUC) of the receiver Operatic Characteristic (ROC) function to evaluate the performance of the proposed FCTBN model. We also compare the performance of the FCTBN with two existing methods from {the} literature including unsupervised MTBN \cite{faruqui2018mining} and LRMCL\cite{alaeddini_mining_2017}. 
{The step-by-step procedure of training and testing of the comparing algorithms is provided in appendix A}.
For the comparisons, considering the patients' existing MCC in the base year, which can be any combination of {the} 5 MCC including no condition, we use each of the comparing methods to predict the future combination of conditions for the next 5 years.

Table \ref{Table:CTBN_ROC} illustrates the AUC performance of the comparing methods for each of the five conditions (presented in the columns) for 2 to 5 years from the baseline (presented in the rows). As can be seen from the table, the proposed functional CTBN generally provides better accuracy compared to MTBN and LRMCL for 4 out or the 5 conditions (Depression, Substance Abuse, PTSD, and Backpain) over both short and long term predictions (2-5 years). However, it shows less predictive power compared to MTBN for forecasting TBI. One justification for the lower prediction accuracy for TBI may be the distinct temporal behavior of TBI occurrences. At the patient level, TBI occurrence is generally a more singular event with chronic clinical ramifications that are coded separately in the electronic medical record. Meanwhile, the performance gap improves for the longer predictions, i.e., years 4 and 5, as it captures the temporal pattern of staying in the TBI state more effectively.

It may also be worth noting that the predictive performance of the functional CTBN as shown in Table \ref{Table:CTBN_ROC} is based on the model trained on risk factors with reduced dimension (1st {principal} component of the risk factors) to improve the computational time, while the MTBN and LRMCL take advantage of the original risk factors. Therefore, we believe training the functional CTBN model with the complete risk factors may further enhance its predictive performance (with the trade-off of increasing the computational time). 

We believe the improved performance of the functional CTBN is partly because of the proposed adaptive group regularization-based learning framework for structure and parameter learning. Specifically, we believe the use of group regularization, in the way proposed in the manuscript, results in fewer yet more informative connections, which improve the training and querying (inference) process. We have shown this concept in one of our related research works \cite{faruqui_2020}. We also believe allowing the model to contain loops (bidirectional connections) helps better capturing the (reinforcing) dynamics of the multiple chronic conditions. Additionally, we think explicit capturing of the time can positively affect the learning of parameters of the model, even though the prediction task is actually on discrete time stamps, i.e., years.

\vspace{-1.5em}
\subsection{Trajectory Analysis}
\label{Subsection:Trajectory_Analysis}
\begin{figure}[H]
    \centering
    \includegraphics[scale = 0.44]{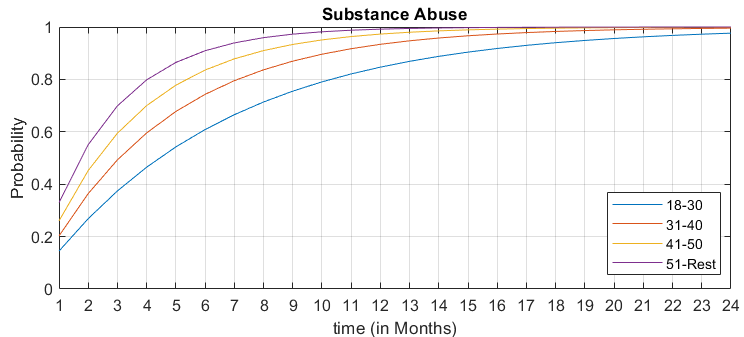}
    \caption{The risk trajectory of developing Substance Abuse disorder over time for patients of different age groups who are diagnosed with TBI and PTSD at baseline.}
    \label{Figure:Trajectory_Case_1}
\end{figure}

An interesting feature of the proposed functional CTBN is the trajectory analysis of state variables (MCC conditions). Here, we demonstrate two cases of MCC trajectory analyses for different preexisting (parent) conditions and age groups (exogenous variables). In the first case, we investigate the effect of age (groups) on the trajectory of Substance Abuse given TBI and PTSD as the preexisting conditions. Figure \ref{Figure:Trajectory_Case_1} shows the most probable trajectory for the emergence of substance abuse for different age groups for the next 24 months given TBI and PTSD in the base month. It can be seen that the \textit{above 51} age group is more prone to be diagnosed with substance abuse than younger age groups, with the probability of developing substance abuse for this age group goes above 80\% just after four months. Whereas, the \textit{18-30} age group reaches 80\% after 10 months, the \textit{31-40} age group passes 80\% mark after 7 months, and the \textit{41-50} group meets the 80\% mark after 5 months. 

This is on par with findings in the medical literature. In a study with service members with mild TBI Miller et al. \cite{miller_2013} found an increased risk for addiction-related disorders including alcohol and nicotine. In a separate study with 6,824 military personnel, Adams et al. \cite{adams_2016} conducted a path based analysis to examine the association of binge alcohol drinking with TBI and PTSD. They found almost 70\% of the total effect of TBI on binge drinking was from the direct path effect, and only 30\% represented the indirect effect through PTSD. Graham et al. \cite{graham_2008} found a decrease in substance abuse post-TBI in a younger age group, likely motivated by significant influences on lifestyle choices and functional status given proper support from VA.

\begin{figure}[H]
    \centering
    \includegraphics[scale = 0.47]{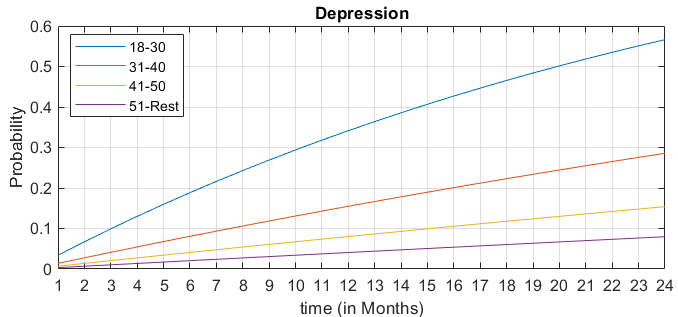}
    \caption{The risk trajectory of developing Depression over time for patients of different age groups who are pre-diagnosed with PTSD.}
    \label{Figure:Trajectory_Case_2}
    \vspace{-1.0em}
\end{figure}

In the second case, we investigate the effect of age (groups) on the trajectory of depression given PTSD as the existing (prior) conditions. Figure \ref{Figure:Trajectory_Case_2} shows the most probable trajectory for the emergence of depression for different age groups for the next 24 months given PTSD in the base month.  
As can be seen from figure \ref{Figure:Trajectory_Case_2} the probability of developing depression after PTSD increases (almost) linearly over time, but with a different slope for different age groups. Unlike the first case, here, the younger patients, i.e., \textit{18-30} age group, are more like to develop depression compared to the other age groups. As the (blue) trajectory line in figure \ref{Figure:Trajectory_Case_2} shows, the \textit{18-30} age group trajectory has a considerably high slope reaching a risk of 50\% after 20 months. Meanwhile, the slope of the trajectory line reduces for older age groups, i.e., the purple line in the figure shows a marginal increase in the risk of depression for patients aged 51 and older. These differences in age group findings may reflect variability in clinical screening approaches, provider biases, and differences in clinical priorities by these patient populations, resulting in increased or decreased likelihood of getting diagnosed with these conditions. For example, younger age group veterans have been undergoing a widespread national screening program to identify PTSD and to establish treatment and follow-up, which would likely lead to the additional diagnosis of depression, a known comorbid condition. 

The medical literature also supports this result.
Lippa et al. \cite{lippa_2015} used factor analysis to identify patterns of comorbidity in a sample of 255 previously deployed Post-9/11 service members and veterans who participated in a structured clinical interview. They found that over 90\% of the patients had psychiatric conditions, and approximately half had three or more conditions. They also identified four clinically relevant psychiatric and behavioral factors, including deployment trauma factor, somatic factor, anxiety factor, and substance abuse factor, which account for 76.9\% of the variance in the data. They concluded that depression, PTSD, and a history of military mild TBI could comprise a harmful combination associated with a high risk for substantial disability. In a separate study, Duncan et al. \cite{duncan_2007} found 36\% of depressed patients screened positive for PTSD. Kobayashi et al. \cite{kobayashi_2007} found that younger and middle‐aged VA primary care patients had more severe PTSD symptoms compared to older patients; however, it was impossible to disentangle the effects of age from a passage of time, limiting interpretation of the findings. On a separate study, on average, women reported age of onset of the depression of 23 –24 years. About half the women were experiencing a mild-to-moderate depressive episode, with 47\% experiencing a severe depressive episode \cite{green_2006}.
Such information can help the medical practitioners develop more individualized plans to prevent the emergence of new chronic conditions according to the patient's specific risk factors and prior conditions. 

\vspace{-1.0em}
\section{Conclusion}
\label{Section: Conclusion}
In this paper, we propose a functional continuous time Bayesian network with conditional dependencies represented by regularized Poisson Regression that can be used to learn both the structure and parameters of the network by solving a non-smooth convex optimization problem. While most Bayesian model structures are sensitive to time granularity since the proposed functional CTBN model can model finite-state continuous time Markov processes over a set of factored states at various time granularity. The model incorporates the functional dependencies among random variables in the Bayesian network structure. It allows for extracting the probability distribution of various combinations of events at different times with respect to any predetermined values of exogenous variables. The model also utilizes an adaptive group regularization method to learn a sparse representation of the system. For the case study, we have used the proposed  CTBN to model the complex temporal relationship among multiple chronic conditions with respect to patient-level risk factors based on a dataset from the Department of Veterans Affairs. 

The proposed model provides a considerable improvement in prediction performance in comparison to Multilevel Temporal Bayesian networks and Latent Regression Markov Mixture Clustering (LRMCL). It also effectively characterizes the trajectory of a medical condition over time when given a different set of starting conditions and risk factors.  
This approach allows for the personalization of the predictive model and therefore has both population and patient-level applications. This modeling approach will inform the clinician about the emergence trajectory of MCC over time and the significant risk factors affecting the trajectory and help guide clinical care to prevent or delay the onset of new chronic conditions.

\bibliographystyle{IEEEtran}

\EOD
\end{document}